\documentclass[times, twoside]{zHenriquesLab-StyleBioRxiv}
\usepackage{blindtext}
\usepackage{lipsum}
\leadauthor{Seungho Choe} 

\begin{document}

% insert title
\title{Teacher-Student Model for Detecting and Classifying Mitosis in the MIDOG 2025 Challenge}
% insert title for footer
\shorttitle{Teacher-Student Model in MIDOG25 Challenge}

% Use letters for affiliations, numbers to show equal authorship (if applicable) and to indicate the corresponding author
\author[1]{Seungho Choe}
\author[1]{Xiaoli Qin}
\author[1]{Abubakr Shafique}
\author[1]{Amanda Dy}
\author[2,3]{Susan Done}
\author[1]{Dimitrios Androutsos}
\author[1,4]{April Khademi}

\affil[1]{Image Analysis in Medicine Lab (IAMLAB), Electrical, Computer and Biomedical Engineering, Toronto Metropolitan University, Toronto, ON, Canada}
\affil[2]{Princess Margaret Cancer Centre, University Health Network, Toronto, ON, Canada}
\affil[3]{Department of Laboratory Medicine and Pathobiology, University of Toronto, Toronto, ON, Canada}
\affil[4]{Vector Institute of Artificial Intelligence, Toronto, ON, Canada}

\maketitle

%TC:break Abstract
%the command above serves to have a word count for the abstract
\begin{abstract}

Counting mitotic figures is time-intensive for pathologists and leads to inter-observer variability. Artificial intelligence (AI) promises a solution by automatically detecting mitotic figures while maintaining decision consistency. However, AI tools are susceptible to domain shift, where a significant drop in performance can occur due to differences in the training and testing sets, including morphological diversity between organs, species, and variations in staining protocols. Furthermore, the number of mitoses is much less than the count of normal nuclei, which introduces severely imbalanced data for the detection task. 
In this work, we formulate mitosis detection as a pixel-level segmentation and propose a teacher-student model that simultaneously addresses mitosis detection (Track 1) and atypical mitosis classification (Track 2).
Our method is based on a UNet segmentation backbone integrates domain generalization modules, namely contrastive representation learning and domain-adversarial training.
%, and multi-scale CNN classifier head for further classification task. 
A teacher–student strategy is employed to generate pixel-level pseudo-masks not only for annotated mitoses and hard negatives but also for normal nuclei, thereby enhancing feature discrimination and improving robustness against domain shift. 
For the classification task, we introduce a multi-scale CNN classifier that leverages feature maps from the segmentation model within a multi-task learning paradigm.
On the preliminary test set, the algorithm achieved an F$_1$ score of 0.7660 in track 1 and balanced accuracy of 0.8414 in track 2, demonstrating the effectiveness of integrating segmentation-based detection and classification into a unified framework for robust mitosis analysis.
%In this work, we propose a semi-supervised mitosis detection algorithm, where a UNet-based segmentation model is utilized to generates a pixel-level pseudo-mask for unannotated mitosis and normal nuclei as well as hard negatives based on annotations. On the preliminary test set, the algorithm recorded an F$_1$ score of 0.7660. 
\end {abstract}
%TC:break main
%the command above serves to have a word count for the abstract

%\begin{keywords}
%segmentation | two-stage
%\end{keywords}

\begin{corrauthor}
seungho.choe@torontomu.ca
\end{corrauthor}

\section*{Introduction}
Mitotic count is an important factor in grading of almost all tumours. For example, the Nottingham grading scheme for breast cancer \cite{bloom1957histological} incorporates mitotic activity as one of three main criteria, where a high mitotic count is related to a more aggressive tumour, highlighting its prognostic importance. 
Recent MIDOG challenges \cite{midog2021, midog2022} have emphasized automated mitotic counting and highlighted intra- and inter-domain generalization as key challenges. Through these experiences, we recognized two main challenges: limited annotation and domain shift. 
First, the lack of well-annotated mitotic figures leads to a weak performance of using supervised learning methods. Also, recent studies have shown that self-supervised approaches are promising for learning representations from limited labeled data, but there is still a challenge in effectively capturing task-specific features in histopathology images~\cite{stacke2020measuring, sohn2020fixmatch}. 
Second, domain shift is another challenge that arises from staining protocols and scanners, and morphological appearances vary across organs and species \cite{stacke2020measuring}, making it harder for AI to achieve cross-domain generalization. Third, automated mitosis detection systems are particularly prone to false positives due to the abundance of non-mitotic nuclei that closely resemble mitotic figures \cite{tellez2018whole, fernandez2022challenging}. 
To address these challenges, recent datasets such as MIDOG++ \cite{aubreville2023comprehensive} and MITOS WSI \cite{aubreville2020completely, bertram2019large} provide multi-domain annotations.
In this study, we propose a unified teacher–student framework that tackles both detection and classification of mitotic figures. For mitosis segmentation (Track~1), the framework generates pixel-level pseudo-masks guided by frozen teacher module, while domain generalization modules--incorporating attention, contrastive learning, and adversarial training--improve stain robustness and reduce false positives. For atypical mitosis classification (Track~2), we extend the framework with a multi-task learning where multi-scale CNN classifier head operating on segmentation feature maps in a multi-task learning setting.

\section*{Materials and Methods}

\begin{figure*}
\centering
\includegraphics[width=17.8cm]{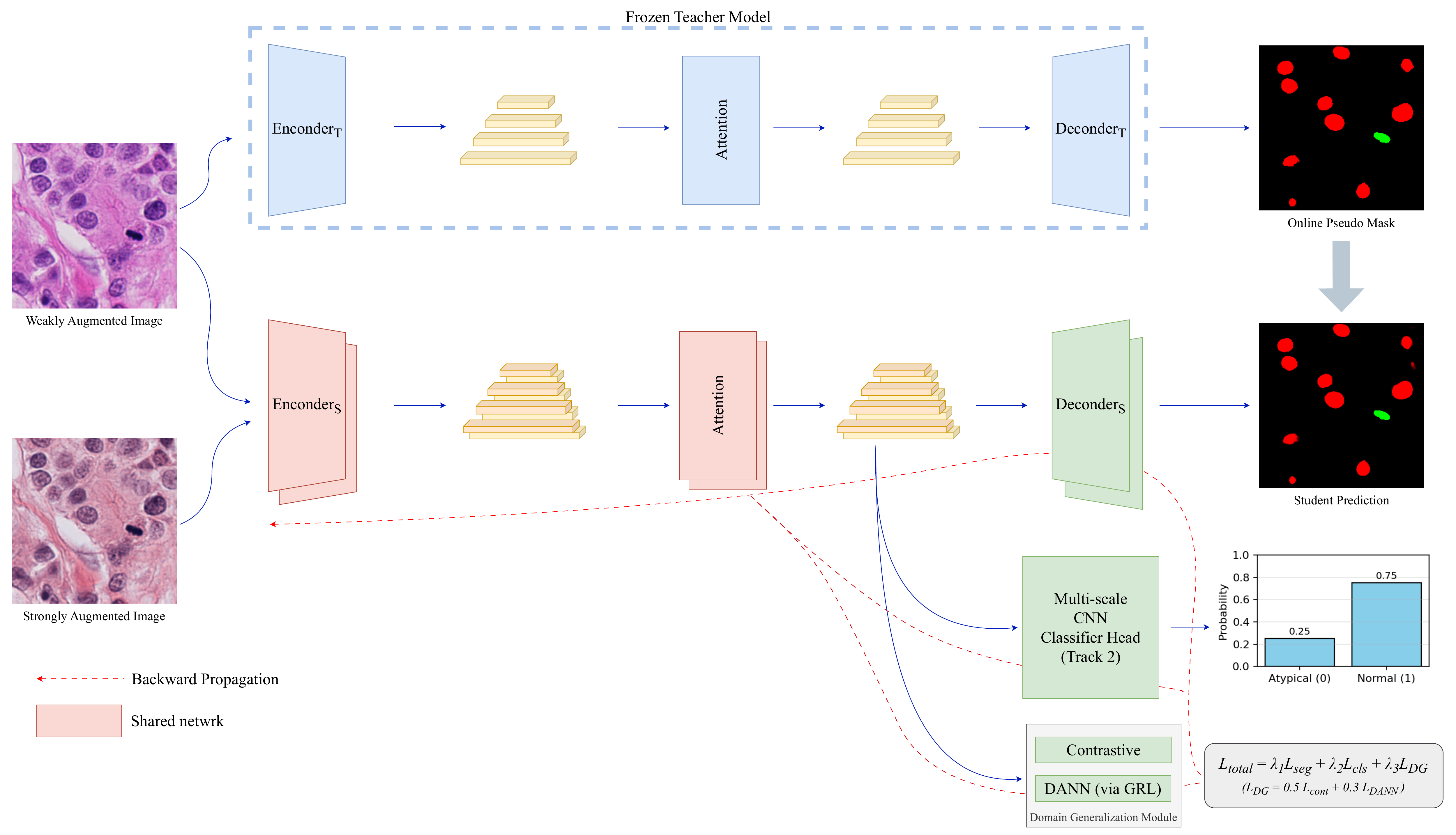}
\caption{Overall pipeline of the proposed semi-supervised framework. The model employs a UNet backbone with attention, contrastive, and domain-adversarial modules for domain generalization. A frozen teacher provides online pseudo-masks that supervise the student decoder via a semi-supervised loss. In addition, a multi-scale classifier head processes encoder feature maps for atypical mitotic figure classification. (Track~2 only)
}
\label{fig:pipeline}
\end{figure*}

\subsection*{Dataset}
We utilized publicly available pathology datasets for both segmentation (Track~1) and atypical mitosis classification (Track~2).  

\paragraph{Track 1.} 
For mitosis segmentation, we employed five datasets: PanNuke \cite{gamper2019pannuke}, TUPAC16 \cite{veta2019predicting, bertram2020pathologist}, MIDOG++ \cite{aubreville2023comprehensive}, and MITOS\_WSI (CMC \cite{aubreville2020completely}, CCMCT \cite{bertram2019large}).  
PanNuke, which provides pixel-level nuclei masks, was used exclusively during the first ten epochs as a warm-up stage to encourage nuclei awareness and stain-robust feature learning; it was excluded from validation and testing.  
The remaining datasets contain mitosis and hard-negative annotations across multiple domains and were used for semi-supervised training, validation, and evaluation.  
Table~\ref{tab:datasets_t1} summarizes their properties.

\begin{table}[h]
\centering
\scriptsize
\caption{Summary of datasets for track 1 used in this study.}
\label{tab:datasets_t1}
\begin{tabular}{lcccc}
\hline
Dataset                   & Data unit      & Annotation       & \# Cases & \# Domains \\ \hline
PanNuke \cite{gamper2019pannuke}           & 256$\times$256 & Pixel-level mask & 2656     & 19         \\
TUPAC16 \cite{veta2019predicting, bertram2020pathologist}           & ROIs           & Centroid         & 73       & 1          \\
MIDOG++ \cite{aubreville2023comprehensive}           & ROIs           & Bounding box     & 503      & 7          \\
MITOS-CMC \cite{aubreville2020completely}   & WSIs           & Bounding box     & 21       & 1          \\
MITOS-CCMCT \cite{bertram2019large} & WSIs           & Bounding box     & 32       & 1          \\
\hline
\end{tabular}
\end{table}
\paragraph{Track 2.} 
For atypical mitosis classification, three datasets were employed: AMi-Br \cite{bertram2025histologic}, the MIDOG 2025 atypical training set \cite{weiss_2025_15188326}, and Octopath \cite{shen_2025_16107743}.  
AMi-Br provides subtype annotations for TUPAC16 \cite{veta2019predicting} and the MIDOG 2021 challenge dataset \cite{midog2021}. The MIDOG 2025 atypical training set offers subtype annotations for the MIDOG++ dataset \cite{aubreville2023comprehensive}. Since MIDOG 2021 is a subset of MIDOG++, overlapping cases were excluded.  
The Octopath dataset contains 3,012 annotated images; however, 1,255 images were either non-mitotic or of unknown class and were therefore excluded.  
After filtering, a total of 15,668 images were used, comprising 12,137 normal mitotic figures and 3,551 atypical mitotic figures.  
Table~\ref{tab:datasets_t2} summarizes the dataset composition.

\begin{table}[h]
\centering
\small
\caption{Summary of datasets for track 2 used in this study.}
\label{tab:datasets_t2}
\begin{tabular}{lcccc}
\hline
Dataset                             & Image size     & \# Total   & \# Normal & \# Atypical \\ \hline
AMi-Br \cite{bertram2025histologic} & 128$\times$128 & 3,720       & 2,888     & 832         \\
MIDOG \cite{weiss_2025_15188326}  & 128$\times$128 & 11,939      & 10,191   & 1,748          \\
Octopath \cite{shen_2025_16107743}  & 64$\times$64   & 3,012       & 379       & 1,378          \\
Total                               &    -           & 15,668      & 12,137    & 3,551          \\

\hline
\end{tabular}
\end{table}

\subsection*{Segmentation Framework}
Figure~\ref{fig:pipeline} illustrates the overall architecture of our proposed semi-supervised segmentation framework. It consists of a UNet \cite{ronneberger2015u} backbone with domain generalization modules, and a frozen teacher module to guide the student model. The UNet is used as the backbone network, which consists of an encoder-decoder architecture. Two augmented images--weak and strong augmentations--are fed into the same segmentation network. Encoder features are passed to domain generalization modules, while the decoder outputs four-class masks (background, normal nuclei, mitosis, and hard negatives). CBAM-style attention \cite{woo2018cbam} enhances spatial and channel-wise focus.
\subsection*{Frozen Teacher Module}
The frozen teacher module generates pseudo-masks online during training. 
For Track 1, we employed the PanNuke dataset to pre-train the model, enabling it to learn the morphological appearance of nuclei. In contrast, for Track 2, PanNuke was not directly applicable; therefore, pseudo-masks were derived using classical morphological operations. Specifically, we extracted the H channel from Macenko stain normalization \cite{macenko2009method}, applied Gaussian blurring, followed by adaptive Otsu thresholding \cite{otsu1975threshold}, and morphological opening to refine nuclei segmentation. 
During training, the student model learns discriminative features of mitotic figures, while the frozen teacher synchronizes with the student model whenever a new best validation score is achieved. In this scheme, the teacher network does not undergo gradient-based updates; instead, its weights are replaced with those of the student at validation checkpoints. The segmentation loss is subsequently computed between the student’s predictions and the teacher-generated pseudo-masks, guiding the student toward domain-robust feature representations.
\subsection*{Domain Generalization Modules}
Domain generalization modules are used to make the model more robust to varying colour distributions. A contrastive learning \cite{chen2020simple} and domain adversarial neural network (DANN) \cite{ganin2016domain} are utilized and the respective losses are computed using two feature maps from two images that differ in the augmentation applied.  The contrastive loss enforces consistency between augmented views, while DANN discourages domain-specific features via gradient reversal layers (GRL) \cite{ganin2015unsupervised}.
\subsection*{Multi-scale CNN Classifier}
The proposed classifier head is designed to refine predictions based on the features extracted from segmentation backbone. Specifically, the classifier operates on multi-scale feature maps extracted from the encoder. One intermediate feature map is passed through a ResNet-152 \cite{he2016deep} refinement block to capture localized discriminative patterns, while the remaining feature scales are globally average pooled and linearly projected to a shared embedding space. These representations are subsequently fused via concatenation, followed by a squeeze-and-excitation gate \cite{hu2018squeeze} that adaptively recalibrates channel-wise responses. Finally, a fully connected layer outputs a binary logit indicating the presence or absence of mitotic figures. This architecture enables effective reuse of encoder features, balances local and contextual information, and provides a computationally efficient yet expressive alternative to employing a deep standalone classifier.

\subsection*{Loss Functions}
The overall training objective combined (1) a semi-supervised segmentation loss, 
(2) a contrastive loss, (3) a domain adversarial loss, and (4) a classification loss in case of track 2. 
The segmentation loss was defined on pixel-level masks, where annotated mitosis and hard negatives were 
combined with teacher-generated pseudo masks for normal nuclei. 
This loss included cross-entropy, focal, and adaptive Dice terms. The adaptive Dice loss is a weighted Dice loss \cite{sudre2017generalised}, where weights are adaptively determined by the class ratios within each batch. 
Additionally, a point-based cross-entropy loss \cite{bearman2016s} was added to encourage higher recall of mitotic figures. 
The overall loss was therefore defined as follows,
\begin{equation}
    \mathcal{L}_{semi}=\mathcal{L}_{CE}+\mathcal{L}_{ADice}+\mathcal{L}_{focal} +\lambda_{1}\mathcal{L}_{pointCE}
\end{equation}
\begin{equation}
    \mathcal{L}_{DG}=0.5\mathcal{L}_{cont} + 0.3\mathcal{L}_{Domain}    
\end{equation}
\begin{equation}
    \mathcal{L}_{cls}=0.5(\mathcal{L}_{BCEpos}+\mathcal{L}_{BCEneg}) + 0.25(\mathcal{L}_{Fpos}+\mathcal{L}_{Fneg})
\end{equation}
\begin{equation}
    \mathcal{L}_{total} = \mathcal{L}_{semi}+\mathcal{L}_{DG} + \lambda_{2}\mathcal{L}_{cls},
\end{equation}
% ADD loss fuctions
where $\lambda_{1}$ is 0.4 and 0.5 for track 1 and track 2 respectively, and $\lambda_{2}$ is 0 and 1 for track 1 and track 2 respectively.
\subsection*{Evaluation}
We used different evaluation metric for each task. The F${_1}$ score as the validation metric for the track 1, and the balanced accuracy for the track 2. Each metrics are defined as follows:
$$ \text{F}_{1}=\dfrac{2\times \text{Precision}\times \text{Recall}}{\text{Precision} + \text{Recall}}$$

\begin{equation}
    BA=\dfrac{1}{2}\bigg(\dfrac{TP}{TP+FN}+\dfrac{TN}{TN+FP}\bigg)
\end{equation}

\section*{Results}
We split the dataset at the patient level allocating 85\% to the training set and 15\% to the test set. 
The training set was further divided, with 80\% used for model training and the remaining 20\% reserved for validation.
In case of WSI datasets, we follow the suggested split scheme in \cite{aubreville2020completely, bertram2019large}. All regions were patched into $512\times512$ tiles with 50\% overlap to avoid missing mitotic figures at the patch borders. 
Data augmentation was designed to generate $256\times256$ inputs while preserving context after rotation. Specifically, we first cropped a $363\times363$ region (approximating $256\sqrt{2}$) at a random location, applied a random rotation, and then performed a center crop to $256\times256$. 
This ensured that the rotated patches retained full coverage without boundary truncation. For stronger perturbations, we applied stain jitter, random blur, and sharpening. 
With 100 total training epochs, we applied early stopping when there is no improvement in the validation metric for 10 consecutive epochs.
We utilized the AdamW optimizer (initial learning rate $4\times10^{-4}$, weight decay $1\times10^{-5}$). The learning rate was gradually increased during the warm-up phase and subsequently decayed with a cosine annealing scheduler to $1\times10^{-6}$. \\
\indent For track 1, we evaluated our proposed method on MIDOG++ and TUPAC16 as in-domain (ID) test. Following the MIDOG 2025 challenge protocol \cite{ammeling_mitosis_2025}, we measured the micro F1-score, counting a prediction as true positive when distance between centroid of prediction and target is less than or equal to $7.5\mu m$. Table~\ref{tab:id_results} outlines the results of the in-domain and preliminary evaluation test. For the in-domain test, we obtained a micro F$_1$-score of 0.7896 on the ID test. Preliminary test set consisted of four domains with 5 cases per each domain. Similar to the MIDOG++ data, each case was cropped tumor region with an area of $2mm^2$. We evaluated our model on the preliminary test set, which resulted in F$_1$-score of 0.7660.
Performance degradation was primarily observed in the second domain.
% use best preliminary results
\begin{table}[h]
\centering
\scriptsize
\caption{Task1: F1 score for mitosis detection}
\label{tab:id_results}
\begin{tabular}{lcccccc}
\hline
          & MIDOG  & Pre. (overall) & D$_1$     & D$_2$    & D$_3$    & D$_4$ \\ \hline
$F_1$     & 0.7896 & 0.7660         & 0.8649    & 0.7319   & 0.7649   & 0.8000   \\
Precision & 0.8932 & 0.8264         & 0.8421    & 0.8783   & 0.7619   & 0.8750   \\
Recall    & 0.7076 & 0.7139         & 0.8889    & 0.6273   & 0.7680   & 0.7368   \\
\hline
\end{tabular}
\end{table}

Similarly, we used AMi-Br, MIDOG 2025, and Octopath datasets for evaluation of the classification task. For the in-domain test, the choice of classification threshold was critical, as inappropriate thresholds could substantially degrade BA.  In our model, the optimal threshold was determined as 0.590, and BA was reported at this operating point. At this threshold, the model achieved a BA of 0.8760, with sensitivity of 0.8627 and specificity of 0.8893 
These results demonstrate that our framework maintains a favorable balance between sensitivity and specificity despite the imbalanced class distribution. 
In the preliminary evaluation, the model achieved a BA of 0.8418 (Table~\ref{tab:t2_result}). 
We observed a slight decrease in BA compared to the in-domain test, primarily due to a significant drop in specificity. 
The increased margin between sensitivity and specificity indicates a tendency toward recall-driven predictions when applied to unseen domains.

\begin{table}[h]
\centering
\caption{Task2: Mitosis and atypia classification performance.}
\label{tab:t2_result}
\begin{tabular}{lcc}
\hline
            & In-domain Test   & Preliminary   \\ \hline
BA          & 0.8760 & 0.8418 \\
Sensitivity & 0.8627 & 0.9155 \\
Specificity & 0.8893 & 0.7682 \\
\hline
\end{tabular}
\end{table}

\section*{Discussion}
In this work, we proposed a student-teacher framework that leverages pseudo-masks for unannotated nuclei to reduce false positives and mitigate the issue of sparse annotations. Our approach achieved a preliminary F$_1$ score of 0.7660 for track 1 and balanced accuracy of 0.8418 for track 2, suggesting that incorporating nuclei pseudo-masks can effectively improves mitosis detection and the multi-task strategy can improve classification accuracy even with limited data. 
Furthermore, the domain generalization modules facilitated the reduction of stain variability by encouraging stain-robust feature learning. 

While these results are promising, this study is limited by the absence of a detailed ablation analysis to quantify the contribution of each component. As future work, we plan to conduct ablation studies and extend our framework to larger multi-organ datasets to further validate its generalizability.

\section*{Bibliography}
\bibliography{literature}
\end{document}